\documentclass[10pt,twocolumn,letterpaper]{article}

\usepackage{iccv}
\usepackage{times}
\usepackage{epsfig}
\usepackage{graphicx}
\usepackage{amsmath}
\usepackage{amssymb}
\usepackage{xcolor}
\usepackage{subfigure}
\usepackage{multirow}
\usepackage{comment}
\usepackage{enumitem}
\newcommand{\HM}[1]{{\color{blue}{[}\textbf{Huizi: #1}{]}}}

\newcommand{\XY}[1]{{\color{cyan}{[}\textbf{Xiaodong: #1}{]}}}

\usepackage[pagebackref=true,breaklinks=true,letterpaper=true,colorlinks,bookmarks=false]{hyperref}

 \iccvfinalcopy 

\def\iccvPaperID{3572} 

\ificcvfinal\fi
\begin{document}

\title{A Delay Metric for Video Object Detection: \protect\\ What Average Precision Fails to Tell}

\author{Huizi Mao\\
Stanford University\\
{\tt\small huizimao@stanford.edu }
\and
Xiaodong Yang\\
NVIDIA \\
{\tt\small xiaodongy@nvidia.com }
\and
William J. Dally \\
Stanford University \& NVIDIA \\
{\tt\small dally@stanford.edu}
}
\maketitle

\begin{abstract}
Average precision (AP) is a widely used metric to evaluate detection accuracy of image and video object detectors. In this paper, we analyze object detection from videos and point out that AP alone is not sufficient to capture the temporal nature of video object detection. To tackle this problem, we propose a comprehensive metric, average delay (AD), to measure and compare detection delay. To facilitate delay evaluation, we carefully select a subset of ImageNet VID, which we name as ImageNet VIDT with an emphasis on complex trajectories. By extensively evaluating a wide range of detectors on VIDT, we show that most methods drastically increase the detection delay but still preserve AP well. In other words, AP is not sensitive enough to reflect the temporal characteristics of a video object detector. Our results suggest that video object detection methods should be additionally evaluated with a delay metric, particularly for latency-critical applications such as autonomous vehicle perception.
\end{abstract}

\section{Introduction}

There is a growing interest in video object detection. Many real-world applications, such as surveillance analysis and autonomous driving, deal with video streams. Several single-image object detection algorithms have been proposed in the past few years~\cite{dai2016r,liu2016ssd,ren2015faster}, but they are compute-intensive to run on a full-resolution video stream. Exploiting temporal information is therefore an important direction to improve the accuracy-cost trade-off~\cite{kang2018t,liu2018mobile,yang2018prernn}.

Prior research suffers the lack of densely annotated video datasets. KITTI~\cite{geiger2013vision} is a dataset targeting at autonomous driving that provides frame-level bounding box annotations. However, it is relatively small compared with other large-scale datasets for training deep neural networks. Since the introduction of object detection from video challenge (VID)~\cite{deng2009imagenet}, more research focus has been drawn into the study of video object detection algorithms. 

There are two general goals of video object detection: improving detection accuracy~\cite{bertasius2018object,feichtenhofer2017detect,kang2018t,zhu2017flow} and reducing computational cost~\cite{chen2018optimizing,mao2018catdet,zhu2017deep}.
Currently, the accuracy of proposed detection algorithms are mostly evaluated with average precision (AP) or mean average precision (mAP) that is the average of APs over all classes~\cite{deng2009imagenet,geiger2013vision,lin2014microsoft}. Video object detection benchmarks like VID also adopt the mAP, where every frame is treated as an individual image for evaluation. However, such an evaluation metric ignores the temporal nature of videos and fails to capture the dynamics of detection results, e.g., a detector that detects the later half occurrences of an instance holds the same mAP as a detector that detects every other frame. As indicated in later experiments, video detectors tend to demonstrate different temporal behaviors compared to their single-image counterparts. 

We introduce average delay (AD), a new detection delay metric. Measuring video object detection delay seems trivial, as the delay can be simply defined as the number of frames from when an object appears to when it is detected. However, to avoid the case where an algorithm trivially detects every bounding box in an image, a false alarm rate constraint is necessary. AD also needs to be designed to be comprehensive like AP, so that the delays at different false alarm rates can be combined. We discuss our design rationale in Section~\ref{sec:metric}.

\begin{figure}[t]
    \centering
    \includegraphics[width=0.5\textwidth]{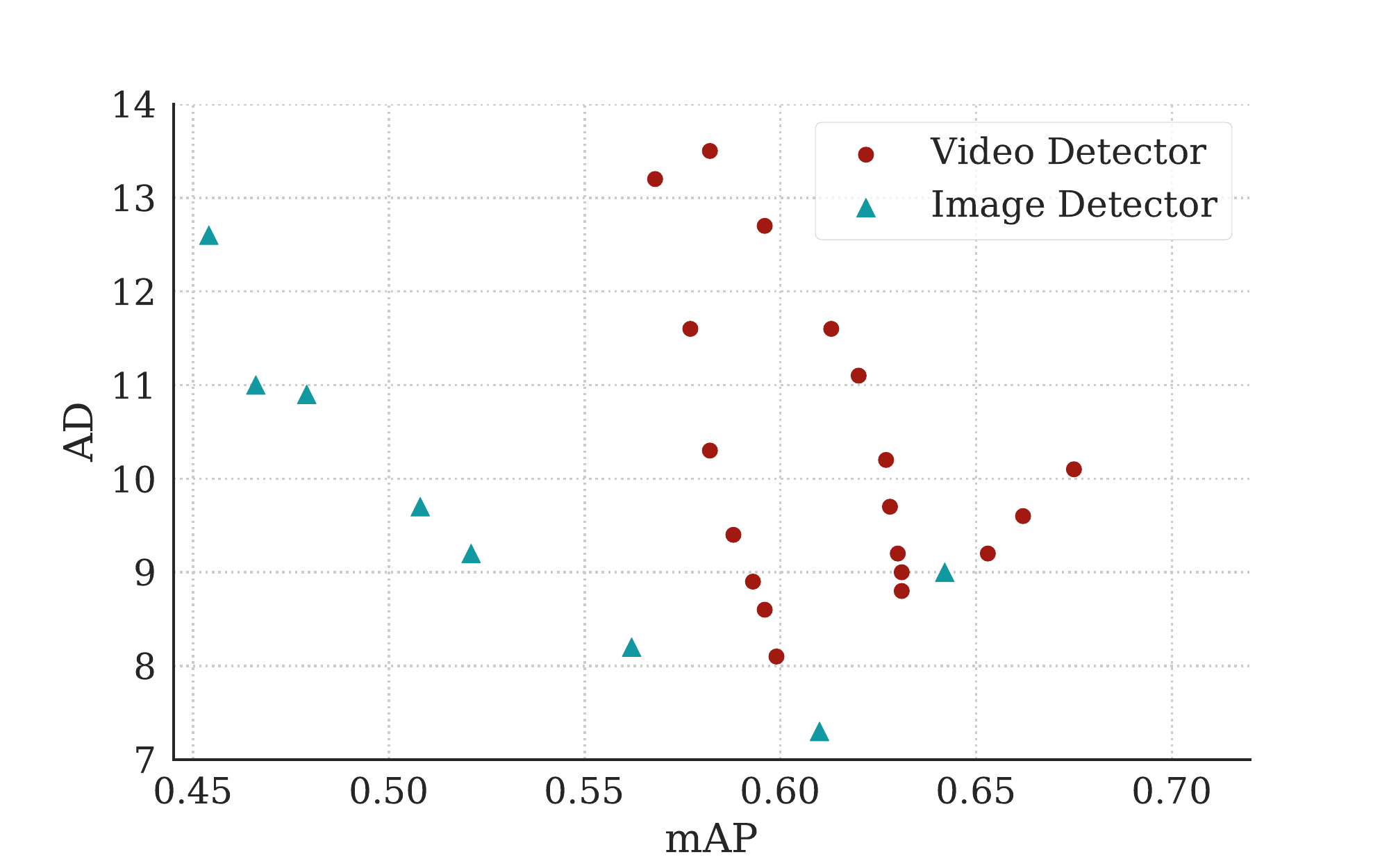}
    \caption{AD does not strongly correlate with mAP. Many algorithms that are specifically designed for video object detection fail to achieve similar AD as the frame-by-frame image detectors, although they may have higher mAP. Image object detectors include R-FCN, Faster R-CNN and RetinaNet. Video object detectors include DFF, FGFA and CaTDet. }
    \label{fig:overall}
\end{figure}

Most video snippets in VID contain fixed numbers of instances (typically only one), which is not suitable for the delay evaluation.
We therefore select a portion of the validation set in VID and name it as VID with multiple tracklets (VIDT). Details of the new VIDT dataset are described in Section~\ref{sec:VIDT}.
With VIDT we then evaluate the AD of a wide range of the recent proposed video detection algorithms in Section~\ref{sec:exp}. A general trend is shown in Figure~\ref{fig:overall}, which indicates that some computation-reducing methods~\cite{mao2018catdet,zhu2017deep} preserve the mAP well but increase the AD. Alternative methods leverage the temporal information to improve detection accuracy but worsen the detection delay~\cite{zhu2017flow}. Our results suggest that video object detection methods should be evaluated with a delay metric, particularly for latency-critical applications such as autonomous vehicle perception.

To our knowledge, this is the first work that brings up and compares delay, a highly critical but usually ignored issue, for the video object detection task. We propose a comprehensive evaluation metric AD to measure and compare video object detection delay\footnote{Code available at \href{https://github.com/RalphMao/VMetrics}{https://github.com/RalphMao/VMetrics}. }. By evaluating a variety of video object detection algorithms, we analyze the key factors for detection delay and provide the guidance for future algorithm design. 



\section{Background}

\subsection{Overview of Video Object Detection}

Video object detection performs a similar task as image object detection, except that the former is carried on a video stream. Densely annotated videos, which are costly to obtain, are typically required to train a video object detector. The ImageNet VID challenge greatly advances the research progress in the field of video object detection, and provides a large frame-by-frame annotated dataset that covers a wide range of scenarios. 

Various methods have since been proposed and evaluated on the VID dataset. The goal of video object detection is to reduce the computational cost or refine the detection results by exploiting the temporal dimension of videos. For instance, deep feature flow (DFF)~\cite{zhu2017deep}, detect or track (DorT)~\cite{luo2018detect}, CaTDET~\cite{mao2018catdet} and saptiotemporal sampling networks~\cite{bertasius2018object} fall into the first category, while T-CNN~\cite{kang2018t}, detect to track (DtoT)~\cite{feichtenhofer2017detect} and LSTM-aided SSD~\cite{liu2018mobile} belong to the second category. These methods are typically variants of the well studied image object detection algorithms such as R-FCN~\cite{dai2016r}, Faster R-CNN~\cite{ren2015faster}, SSD Multibox~\cite{liu2016ssd} and RetinaNet~\cite{lin2017focal}.

As required in the VID challenge, the performance of a video object detector is solely evaluated by mAP, the metric for still image object detection~\cite{deng2009imagenet,everingham2010pascal,lin2014microsoft}. When evaluating mAP, every single frame of a video is treated as an individual image. In such a way, the quality of a detector over the whole video sequence is measured and compared.

\subsection{Low Latency as a Practical Requirement}

Low latency is a common requirement for many video-related applications.
For example, autonomous driving typically requires less than 100ms latency~\cite{lin2018architectural}. 
Detecting an object with minimum delay is desired, and detection after certain time is no longer important.

In previous research, the term latency mostly refers to computational latency only~\cite{berkeleybdd,martinez2010moped}. 
However, we argue that the overall latency equals to \textbf{computational latency} plus \textbf{algorithmic delay}, and the latter is the time taken in a video stream for an algorithm to finally determine the existence of an object. 
Computational latency has been extensively studied in the recent works~\cite{howard2017mobilenets, mao2018towards,zhang2018shufflenet}, while algorithmic delay remains less explored in the object detection field. In other fields like activity detection, there have been efforts to study early detection~\cite{ma2016learning}.

\subsection{Relevant Studies on the Delay Issue}

Quickest change detection (QCD) is a well studied problem in statistical processing. It refers to real-time detection of abrupt changes in the behavior of an observed signal or time series as quickly as possible~\cite{poor2009quickest}. Generally, the delay is measured at a certain constraint of false alarms. 
Under the framework of QCD, Lao et al. targeted firstly the moving object detection problem~\cite{lao2016quickest} and then the general video object detection problem~\cite{lao2019minimum}. Their approach focuses on the first occurrence of an object, thus differs from general video object detection where the following detection also matters.

NAB~\cite{lavin2015evaluating} is a benchmark for real-time anomaly detection in time-series data. The authors pointed out that traditional scoring methods such as precision and recall do not suffice, as they cannot effectively test anomaly detection algorithms for real-time use. To reward early detection, they define anomaly windows. Inside the window, true positive detections are scored by a sigmoid function and  out of the window all detections are ignored.

In the field of video action recognition~\cite{mahasseni2018budget,yang2016multilayer,yang2019step}, early detection has also gained attention~\cite{kong2014discriminative,sadegh2017encouraging}. This task typically requires accumulating enough frames to make a decision. To alleviate this issue, a special loss function was proposed to encourage early detection of an activity~\cite{ma2016learning}.

All of the works above are essentially dealing with the single object or single signal case. CATDet~\cite{mao2018catdet} introduced a delay metric to measure the detection delay for multiple objects. However, the delay is evaluated at a specific precision only to counter false alarms. 

\section{The Average Delay Metric}
\label{sec:metric}

In this section, we present our definition of average delay (AD), the evaluation metric for video object detection delay. 
Our metric is designed to incorporate fairness and comprehensiveness.
\textbf{Fairness}: AD considers the trade-off between false positives and false negatives to avoid the case of reducing delay by detecting many false positives.
\textbf{Comprehensiveness}: AD covers a wide range of operating conditions, analogous to AP.

We first explain the terminology used throughout this paper before delving into the detailed derivation.
An \textit{instance} is a physical object that appears in consecutive frames as a trajectory (or a tracklet). 
An \textit{object} refers to a single occurrence of an instance in a frame. The ground truth of an object includes its bounding box coordinates, class label, and track identity.
A \textit{detection} is the recognition of an object in one frame with bounding box coordinates, class label, and confidence.

\subsection{Delay and Statistical Process of Detection}

The most intuitive definition of delay is the number of frames taken to detect an instance from the frame it appears. Before reasoning on a comprehensive delay metric, we make this simple assumption: a detector detects every object at every frame with the same probability $p$.

\begin{figure}[b]
    \centering
    \includegraphics[width=0.5\textwidth]{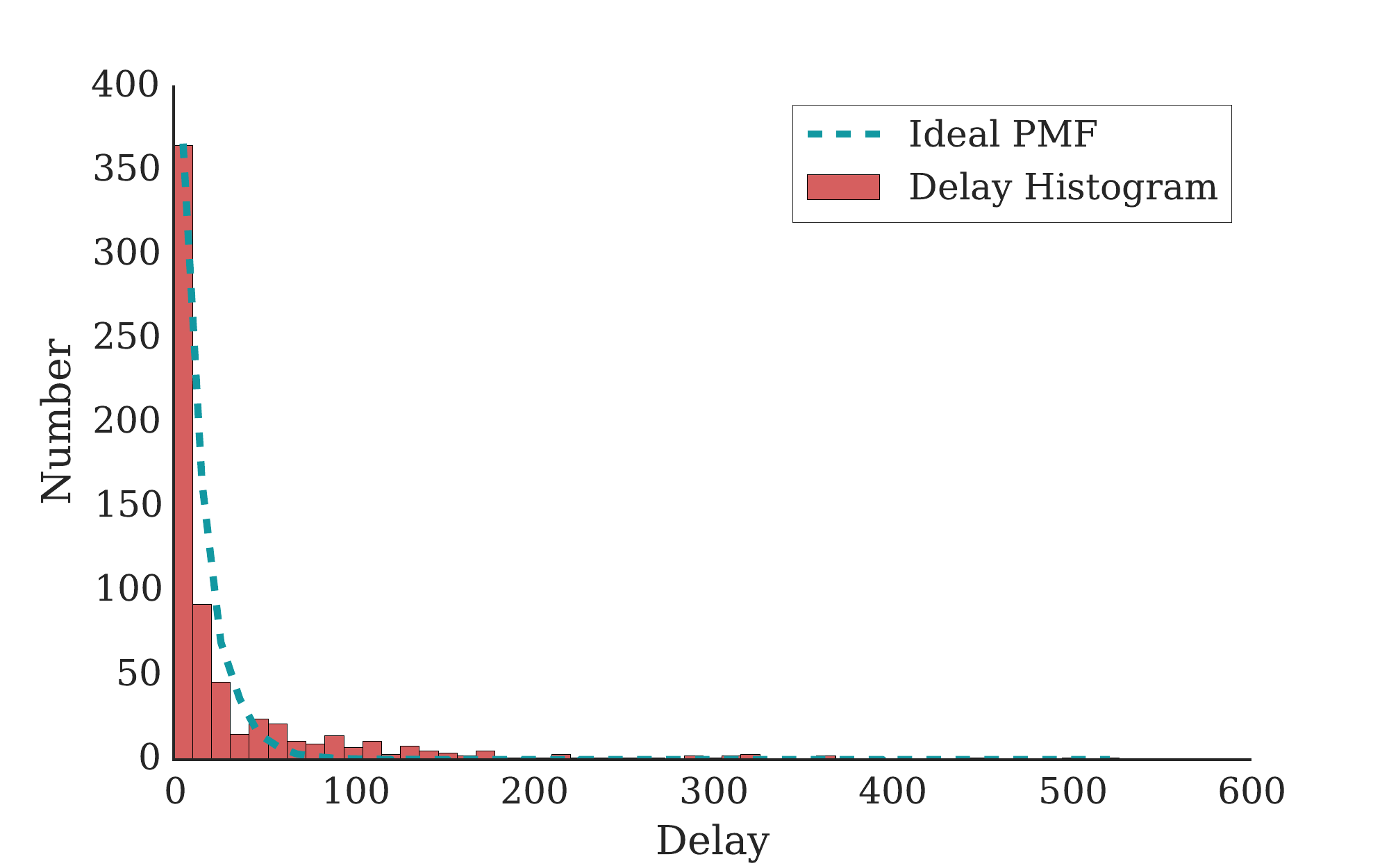}
    \caption{A delay histogram of R-FCN (ResNet-101) on VIDT at a confidence threshold of 0.5. We also show a plot of probability mass function (PMF) under an ideal discrete exponential distribution as the reference to the actual delay dstribution.}
    \label{fig:delay_distribution}
\end{figure}

Under this assumption, the delay $D$ follows the discrete exponential distribution: $D \sim \exp(p)$. 
Figure~\ref{fig:delay_distribution} exemplifies a histogram of the detection delays of R-FCN on VIDT. 
The actual distribution generally resembles the exponential distribution, apart from an anomalous region in the tail. There are substantially more instances than expected with extremely large delays due to existence of ``hard instances''. A detailed discussion about the delay statistics is described in Section~\ref{sec:analysis} and hard examples are given in Figure~\ref{fig:VIDT_hard}.

For discrete exponential distribution,
the expected value follows $E(D)=1/p - 1$. 
Thus, we can measure the quality of a detector through inferring the latent parameter $p$, given multiple observed datapoints ${D_i}$, where $i=1,...,N$.
With maximum likelihood estimation, we find that the maximum likelihood is achieved when the expected value matches the mean of the samples:  $E(D)=\frac{1}{N}\sum_{i = 1}^N D_i = {\bar D}$. So the detection probability $p$ on each frame can be obtained by:
\begin{align}
\label{eq:prob}
    p = \frac{1}{{\bar D} + 1}
\end{align}
As aforementioned, the existence of ``heavy tail'' results in a potential problem when we try to estimate $p$. Different detectors may not be effectively differentiated if the heavy tail dominates the mean value. We thus adopt a simple strategy to clip the delay samples with a constant value $W$, which we name as a \textit{detection window}. This is also a practical consideration, as for most latency-critical tasks a detection no longer matters once it falls out of a time window.
\begin{align}
\label{eq:delay}
\begin{split}
    p &= \frac{1}{{\bar D^*} + 1}, \\
    \bar D^* &= \frac{1}{N} \sum_{i = 1}^N \min(D_i, W).
\end{split}
\end{align}

\subsection{Choice of False Positive Ratio}
It is important to set a threshold for false alarms to ensure fair comparisons. In the previous work~\cite{mao2018catdet}, precision that is defined as number of true positives divided by total detections is selected as a threshold to counter false alarms, as the increased number of false alarms will reduce precision. However, there are undesired outcomes if we set the same precision to compare different detectors.

\begin{figure}[b]
    \centering
    \includegraphics[width=0.5\textwidth]{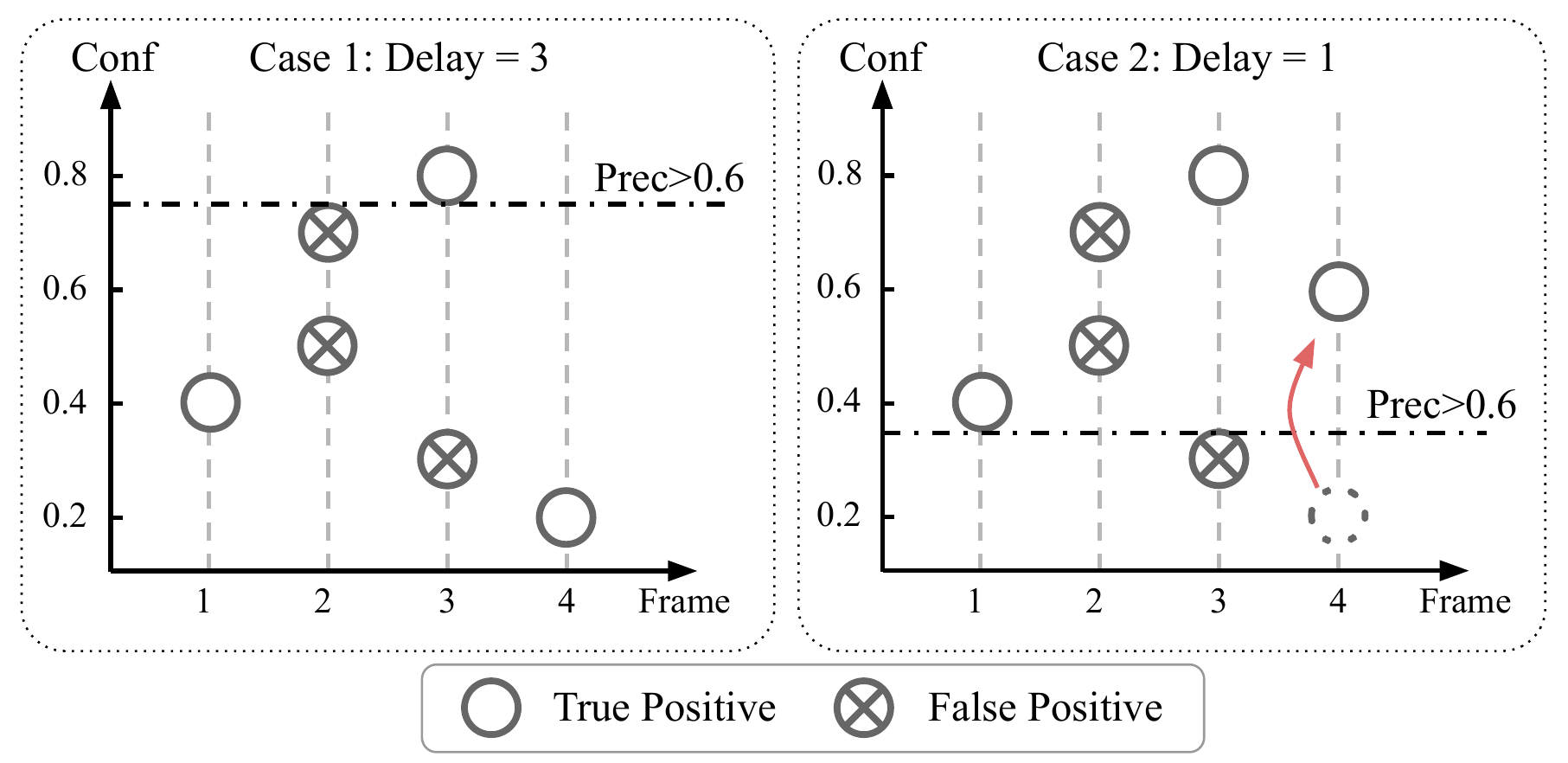}
    \caption{A toy example to illustrate that using precision as the control may lead to undesired behaviors. There is one ground truth instance in frames 1-4. We set the control as Prec $> 0.6$. Due to a more confident true positive at frame 4, case 2 has an unreasonable lower delay than case 1. Setting false positive ratio as the control would avoid this problem.}
    \label{fig:false_alarm}
\end{figure}

\begin{figure*}[h!]
    \centering
    \includegraphics[width=\textwidth]{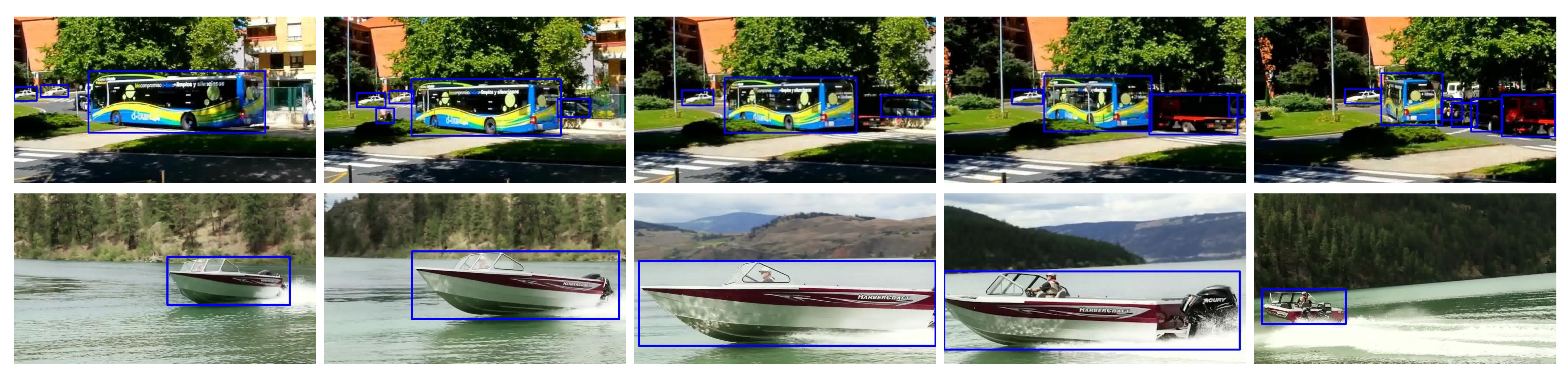}
    \caption{Snippets in the validation set of VID. Top: an ideal video snippet for delay evaluation with multiple instances emerging randomly over space and time. 
    Bottom: an undesired video snippet, in which there is the same instance throughout the time.}
    \label{fig:VID_example}
\end{figure*}

We demonstrate with a toy example in Figure~\ref{fig:false_alarm} to illustrate that setting a precision threshold may cause the measured delay to behave differently from our expectation. 
Suppose the precision threshold is set to 0.6. In case 1, we should set a confidence threshold of 0.75 to meet the precision requirement. In case 2, due to the increased confidence score of the last detection, a confidence threshold of 0.35 is adequate. The resulted detection delay in these two cases are 2 and 0, respectively.
By refining the later detections, the detection delay can be magically improved. Such a behavior counters our intuition that delay should be a matter of the early detections.

As a result, we argue that precision may not be the ideal threshold to counter false alarms. Instead we propose to use false positive (FP) ratio, which is the ratio between false positives and ground truth objects. FP ratio as a threshold is determined only by false positive detections, therefore will not be impacted with more true positives. 

\subsection{A Comprehensive Metric}

The last question comes that how we should comprehensively measure the detection delay of a detector under different false alarm constraints, similar to what AP does. AP is the integral or the arithmetic mean of precisions over different recalls. Analogously, is it a good practice to average detection delays over different false positive ratios?

Consider the real-world scenarios, a detector with zero delay is substantially better than one with 1-frame delay, while a detector with 14-frame delay does not make a significant difference from one with 15-frame delay. However, the arithmetic mean cannot distinguish the two cases. 

We argue that averaging the latent parameter $p$, which represents the probability of detecting an object, would be a better choice. Since $p$ is the reciprocal of ${\bar D} + 1$, it weighs more for a smaller delay. In addition, it is a bounded value between 0 and 1. As a result, we average the inferred $p$ values of a detector under different false positive ratios, and derive the corresponding AD fom the averaged ${\bar p}$. 

We show our definition of the proposed AD in Equation~\ref{eq:ad}. Here $R$ stands for the total number of FP ratios and ${\bar D}^*_r$ is the delay measured by Equation~\ref{eq:delay} at a specific FP ratio $r$. Notice that this definition has a very similar form to the harmonic mean.
\begin{align}
\label{eq:ad}
    AD = \frac{1}{\bar p} -1 = \frac{1}{\frac{1}{R}\sum_r \frac{1}{{\bar D}^*_r+1}}-1
\end{align}
In our following experiments, we set the detection window $W$ to 30 frames and select 6 FP ratios including 0.1, 0.2, 0.4, 0.8, 1.6 and 3.2.


\section{Dataset for Delay Evaluation}
\label{sec:VIDT}

\subsection{Overview}
There are multiple public datasets for object detection, such as KITTI~\cite{geiger2013vision}, ImageNet-VID~\cite{deng2009imagenet}, YouTube-BB~\cite{real2017youtube}, BDD100K~\cite{yu2018bdd100k}, VIRAT~\cite{oh2011large}, etc. However, they suffer from various drawbacks for delay evaluation. KITTI is a relatively small dataset, making it hard to train deep neural networks. Most video snippets in ImageNet VID contain fixed numbers of objects from beginning to end, which leaks strong prior, thus making it unsuitable for delay evaluation. Youtube-BB and BDD100K are both large-scale datasets with rich objects and scenarios, but they are sparsely annotated. VIRAT is a surveillance analysis dataset and has a fixed background.  

An ideal dataset for delay evaluation should (\textbf{\romannumeral 1}) be densely annotated (frame by frame); (\textbf{\romannumeral 2}) have random entry time for each instance (exclude videos with the same objects throughout the time); (\textbf{\romannumeral 3}) have random entry location for each instance (exclude videos with a fixed background and limited entry locations for new objects).
In Figure~\ref{fig:VID_example}, we show examples of ideal and non-ideal snippets in the validation set of ImageNet VID. The ideal snippet has multiple different instances entering the frames randomly over space and time, while in the non-ideal case, the same instance (which is a boat in the example) exists from the very first frame to the last.

\subsection{Introducing VIDT}
We introduce VIDT, a subset of the validation set of VID, to meet the requirements aforementioned. Video snippets in VIDT have at least one instance entering at a non-first frame, which guarantees the randomness of entry time. VIDT largely relies on the annotated track identities in VID. A subtle difference is that in VIDT, once an instance disappears for more than 10 consecutive frames, it is marked as a new instance. One reason is that we do not care about the re-identification ability but only the capability to detect as early as possible. In this way, the number of instances is increased from 555 to 666. 

\begin{figure}[t]
    \centering
    \includegraphics[width=0.53\textwidth]{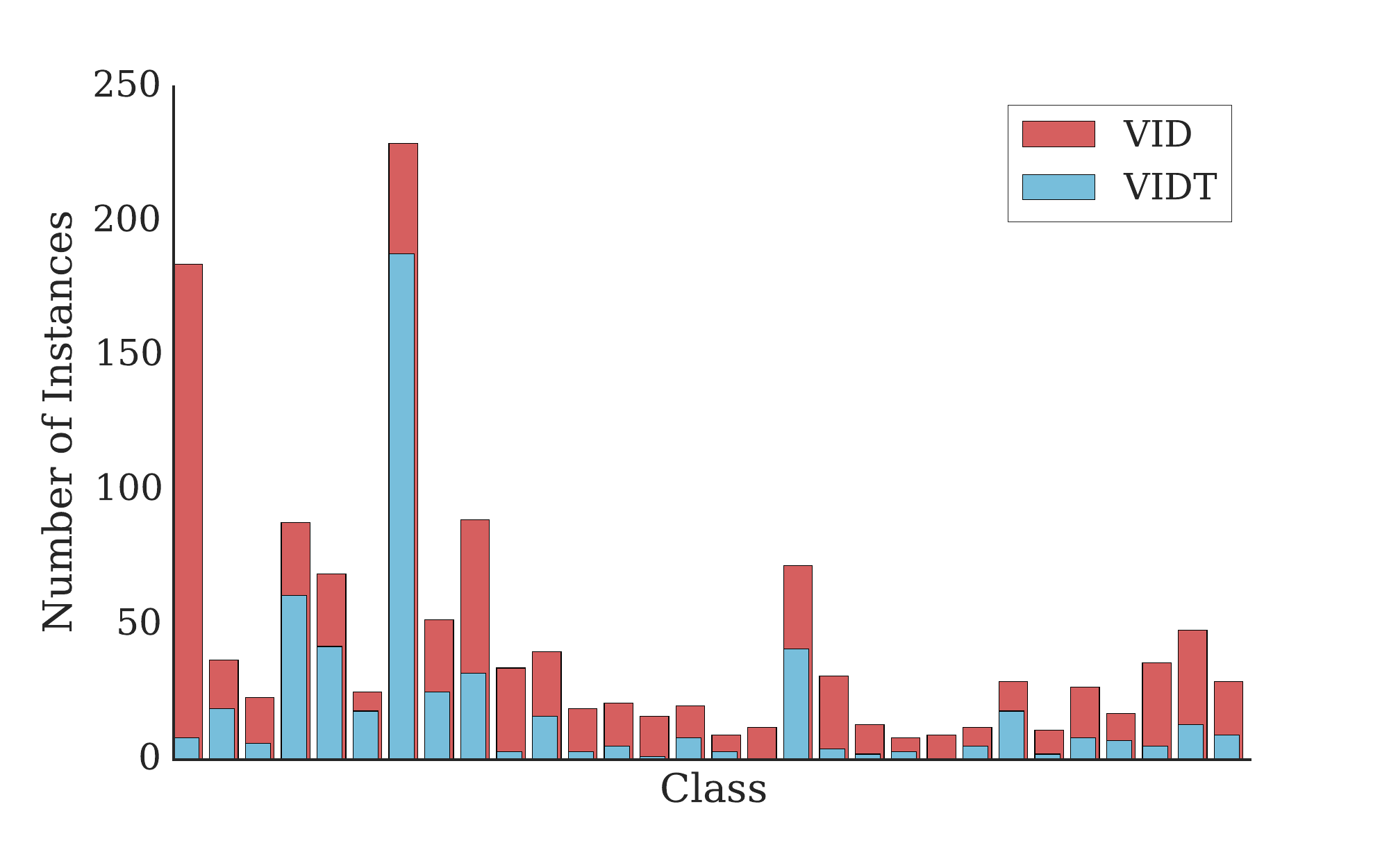}
    \caption{Number of instances per class is highly imbalanced in VID and VIDT. The class ``Car'' has most instances in both datasets. There is no instance of ``Lizard'' and ``Sheep'' in VIDT.}
    \label{fig:class_imba}
\end{figure}

\begin{table}[t]
    \centering
    \begin{tabular}{l|cc|cc}
    Dataset & Snippets & Frames & Instances & Objects \\
    \hline
    \hline
    VIDT & 120 & 53K & 666 & 102K\\
    VID-val & 555 & 176K & 1309 & 274K \\
    VID-train & 3862 & 1122K & 7911 & 1732K \\
    KITTI* & 21 & 8K & 783 & 41K \\
    \end{tabular}
    \vspace{2mm}
    \caption{Statistics of the candidate datasets for delay evaluation. Note that KITTI does not have an official split of train/val.}
    \label{tab:data_statistics}
\end{table}

We report the statistics of VIDT and compare with the original VID and KITTI in Table~\ref{tab:data_statistics}. The ample training data of VID makes it feasible to train deep neural networks. Even though VIDT is smaller than the original validation set of VID, it still has much more frames and objects than KITTI with training and validation sets combined. However, severe class imbalance problem exists in both VID and VIDT as shown in Figure~\ref{fig:class_imba}. Therefore, AD is not measured on each class separately, but instead treats all instances in a class-agnostic way.

\section{Experiments}
\label{sec:exp}

In this section, we demonstrate a common but mostly ignored problem in recent research of video object detection. Many detectors suffer from worse detection delay, even though they are able to preserve or even improve mean Average Precision. 

\subsection{Toy Cases for Metric Comparison}
\label{sec:toy_case}
We design several special cases to show the advantages of our proposed average delay metric against mAP~\cite{everingham2010pascal}, NAB Score~\cite{lavin2015evaluating} and CaTDet Delay~\cite{mao2018catdet}. 
The NAB metric, originally designed for anomaly detection, can be modified to fit in the object detection task. The modification is described in Appendix.

Our comparison of the different metrics is achieved by manipulating the detection output and quantifying the impact on each metric. Retardation measures the \textbf{sensitivity} by suppressing the first few detections of a tracklet. A desirable delay metric should be worsened after retardation. Tail boost measures the \textbf{fairness} by elevating the confidence scores of lately detected objects. A fair delay metric should not be affected by tail boost.
Multiple observations can be drawn from Table~\ref{tab:delayed_detector}.
\begin{itemize}[noitemsep,topsep=0pt]
\item For mAP, retardation makes little impact while tail boost greatly improves the result, which is in accordance to the number of affected detections. 

\item Retardation worsens all three delay metrics. However, if suppressing the low-confident objects only, NAB and CaTDet do not reflect the change, as both of them operate at a single confidence threshold.
In contrast, AD evaluates multiple thresholds, therefore is robust to reflect the effect of retardation.

\item For tail boost, it improves both NAB and CaTDet, while only improves AD negligibly, indicating that AD is better than the other two metrics in term of fairness.
\end{itemize}

\begin{table}[b!]
    \centering
    \begin{tabular}{l|c|cc|c}
         & \multirow{2}{*}{Baseline} & \multicolumn{2}{c|}{Retardation} & \multirow{2}{*}{Tail Boost} \\
         & & Low-Conf & All & \\
\hline
\# of affected & \multirow{2}{*}{0} & \multirow{2}{*}{3076} & \multirow{2}{*}{3616}& \multirow{2}{*}{71781} \\
detections & & & \\
\hline
mAP            &  0.64   &  0.63   & 0.63 & 0.70  \\
\hline
NAB      &  0.29    &  0.29  &  0.17 &   0.31  \\
\hline
CaTDet   &  13.6    &  13.6   & 15.1 &  12.5    \\
\hline
\textbf{AD (Ours)}             &  9.0     &  11.5   & 13.8 &  8.9    \\
    \end{tabular}
    \vspace{2mm}
\caption{Comparison of the different metrics by data manipulation. Baseline is an R-FCN detector with ResNet-101. Retardation makes detection slower by suppressing the first 5 detections of a ground truth instance. In the case low-conf, we only suppress the detections with low confidence, while in the case all, all detections are suppressed regardless of their confidence scores. Tail boost improves the detections that are 20 frames later than the first occurrence of ground truth. Note that for CaTDet and AD, lower numbers indicate better results.}
\label{tab:delayed_detector}
\end{table}

\subsection{Key Frame based Methods}
A range of recent works on video object detection employ the concept of the key frame~\cite{chen2018optimizing,hetang2017impression,luo2018detect,zhu2017deep}. Key frames are sparsely distributed over the whole video sequence and typically require more computational resources than non-key frames. Key frames can be used to improve the detection accuracy or reduce the cost of non-key frames, through exploiting the temporal locality in videos.

\begin{figure}[t]
    \centering
    \includegraphics[width=0.52\textwidth]{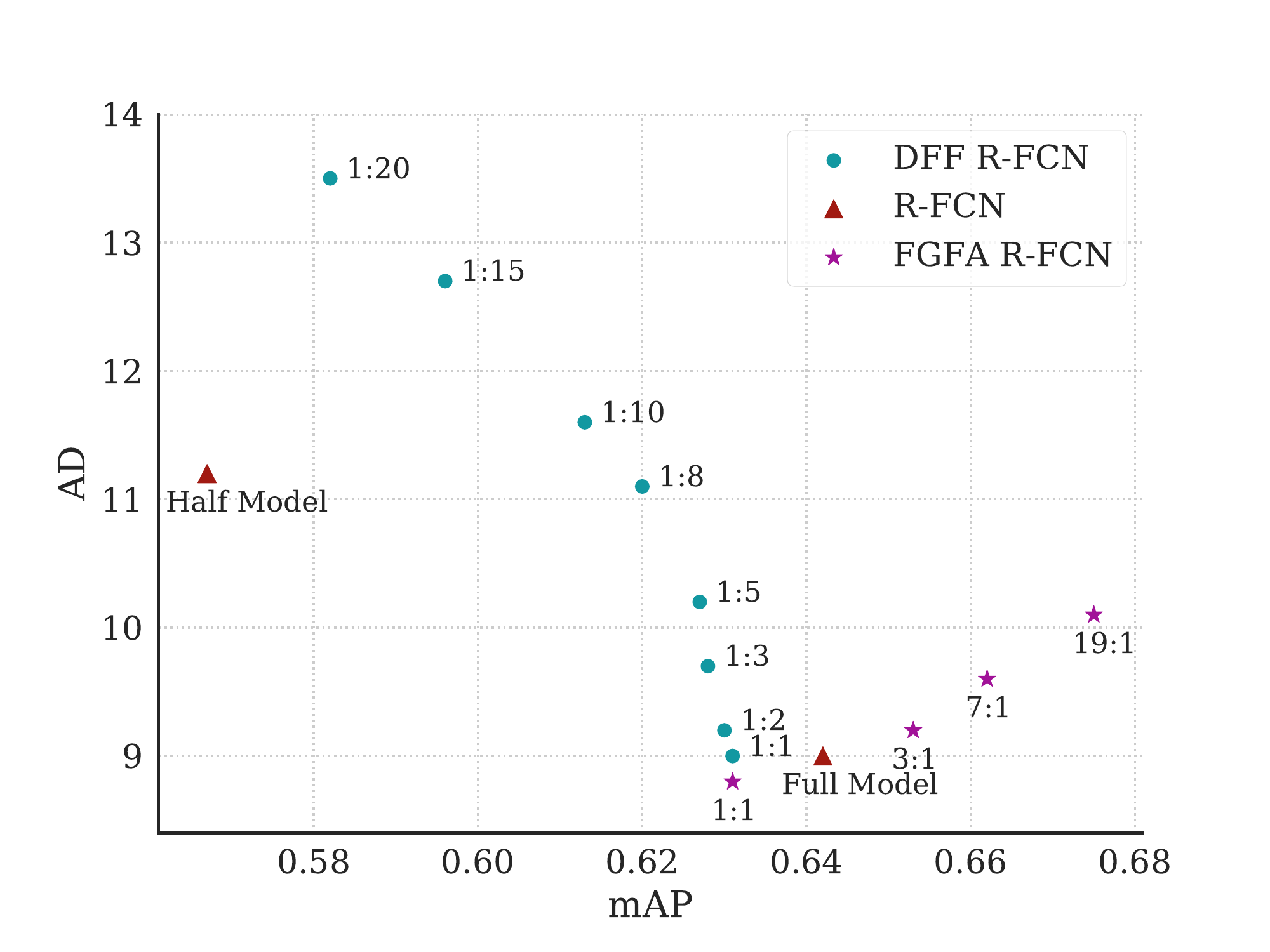}
    \caption{How DFF and FGFA affect mAP and AD. Here $1:N$ refers to 1 key frame in every $N$ frames for DFF. $N:1$ refers to N frames aggregated for FGFA. The full and half models are both frame-by-frame R-FCN models, except that the half model is trained with half number of iterations. All models use ResNet-101 as the backbone.}
    \label{fig:ap_ad_interval}
\end{figure}

We choose deep feature flow (DFF)~\cite{zhu2017deep} as a representative key frame based algorithm. The basic idea is to compute features on key frames and propagate the features with optical flow on non-key frames. We vary the interval of key frames and show the impact on mAP and AD in Figure~\ref{fig:ap_ad_interval}. Two R-FCN models are also reported for comparison. The full model is a standard R-FCN model with ResNet-101, and the half model is in the same architecture but trained with only half number of iterations.

Figure~\ref{fig:ap_ad_interval} shows that DFF tends to worsen AD. For example, the DFF model that adopts a key frame in every 10 frames achieves mAP of 0.613, much higher than the mAP 0.567 of the inferior R-FCN model. However, in term of AD, the DFF model is a bit worse (11.6 vs. 11.2). This indicates that setting sparse key frames leads to the delayed detection of new objects.

\subsection{Feature Aggregation Methods}

Combining features of multiple frames is an effective approach to improve detection accuracy. Recent works in the field include
explicit feature aggregation by temporally adding up features~\cite{bertasius2018object,zhu2017flow} and
implicit feature aggregation via recurrent neural networks~\cite{liu2018mobile}.

We select the flow-guided feature aggregation (FGFA) ~\cite{zhu2017flow} as an example and demonstrate how it may affect detection delay while improving mAP. FGFA aggregates the features of previous frames and solves the spatial mismatches by propagating the features with optical flow. The open-sourced version of FGFA is based on R-FCN, therefore we also compare its mAP and AD in Figure~\ref{fig:ap_ad_interval}. FGFA alone improves mAP from 0.642 to 0.675, meanwhile deteriorates AD from 9.0 to 10.2. We also observe a trend that the more frames aggregated, the better mAP can be obtained but the worse AD is.

\begin{figure}[t]
    \centering
\begin{subfigure}

    \includegraphics[width=0.5\textwidth]{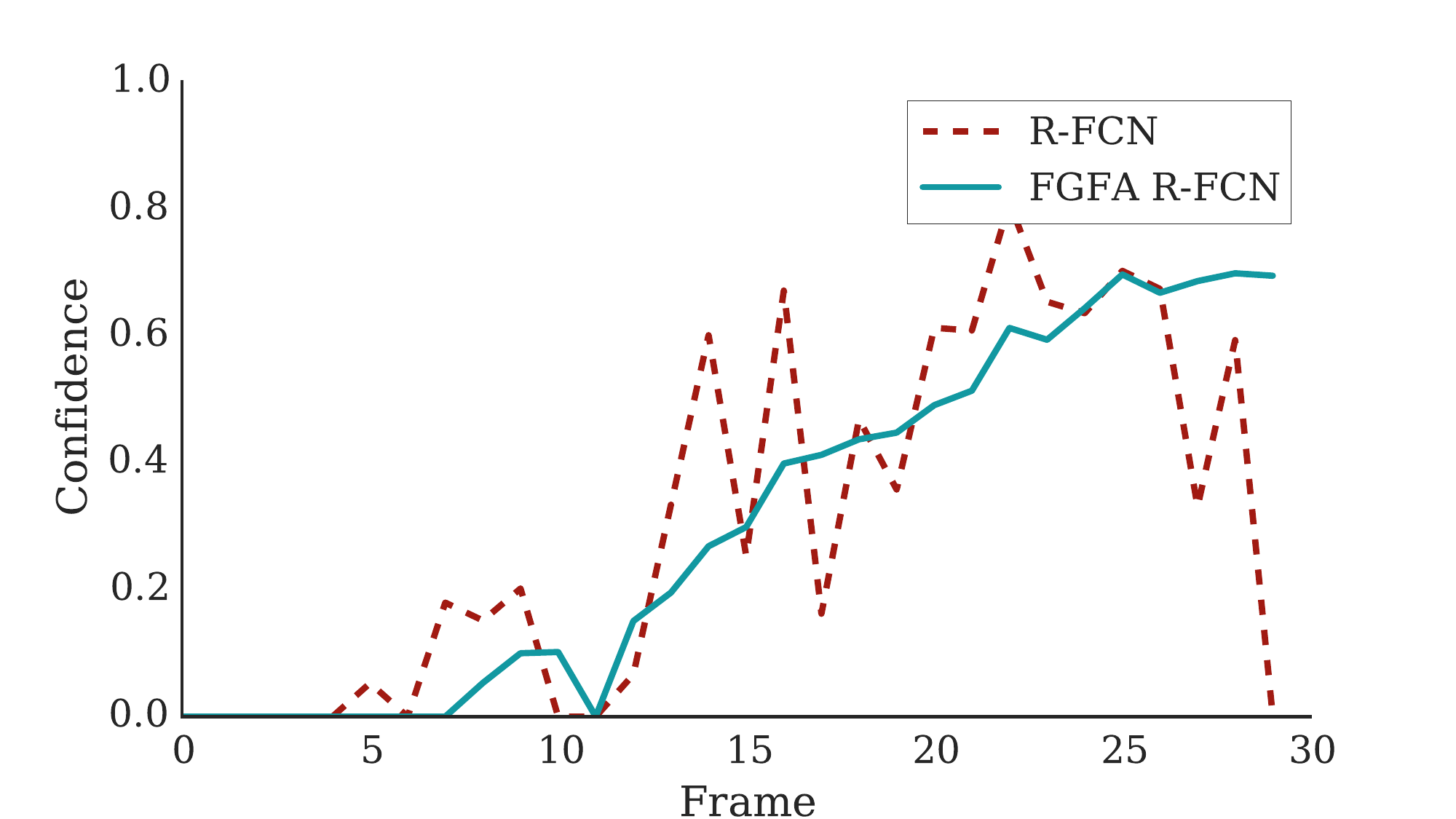}
\end{subfigure}
\begin{subfigure}

    \includegraphics[width=0.5\textwidth]{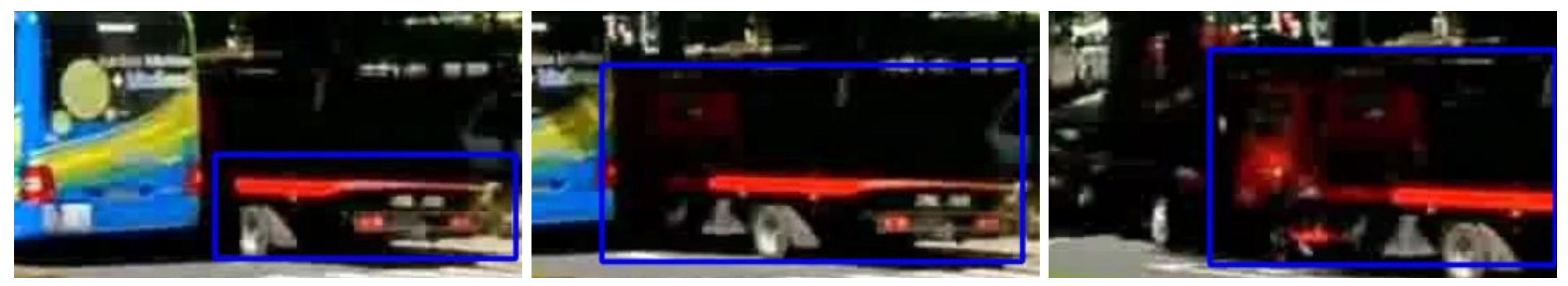}
\end{subfigure}
    \caption{An example illustrates how FGFA causes higher detection delay. The frame-by-frame R-FCN model exhibits large fluctuation of confidence, while the FGFA model tends to slowly build up the confidence over time.}
    \label{fig:fgfa}
\end{figure}

FGFA substantially improves mAP compared with the original R-FCN, but worsens the detection delay. To explain this phenomenon, we select one instance with increased delay and plot the process of being detected in Figure~\ref{fig:fgfa}, which shows the confidence score of the closest detection to the ground truth object. In the case where no detection has an IoU over 50\%, the confidence score is 0. The steady and progressive increasing confidence of FGFA, as shown in the figure, incurs the extra delay to detection, suggesting that for latency-critical tasks it is probably not a good choice to slowly build up the confidence.

\subsection{Cascaded Detectors}
A cascaded detector consists of multiple components and tries to shift the workload from complex ones to simple ones, following specific heuristics. 
Bolukbasi et al.~\cite{bolukbasi2017adaptive} proposed a selective execution model for object recognition problem, which is essentially a cascaded system.
Further works explored the efficacy of cascaded systems in the video object detection task, including scale-time lattice~\cite{chen2018optimizing} and CaTDet~\cite{mao2018catdet}.

\begin{figure}[t]
    \centering
    \includegraphics[width=0.53\textwidth]{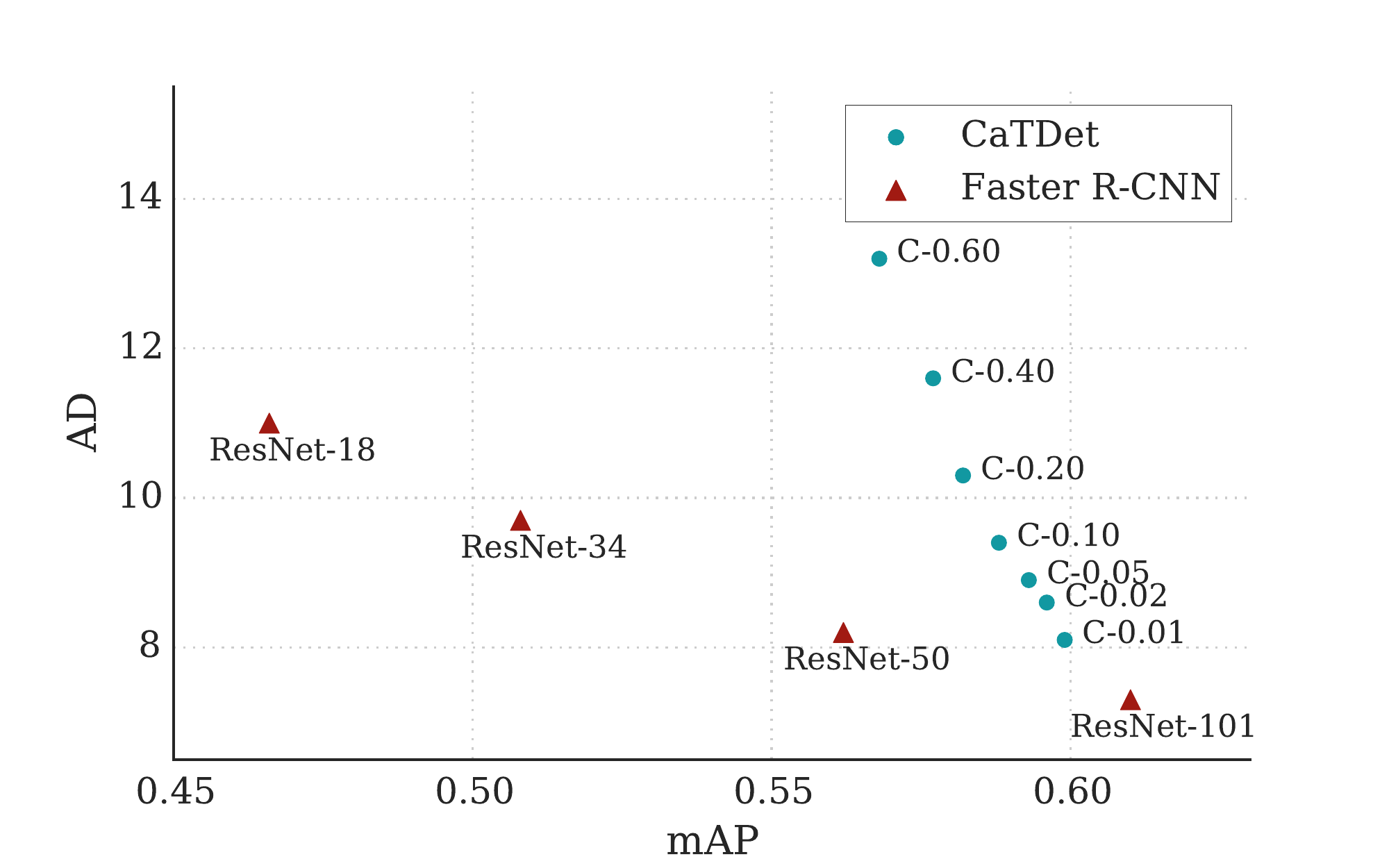}
    \caption{CaTDet preserves mAP well but incurs more AD, compared with Faster R-CNN with smaller models. C-$\alpha$ stands for CaTDet with an intermediate threshold of $\alpha$. Larger $\alpha$ value saves more computation at the cost of more accuracy loss. All CaTDet models are based on Faster R-CNN with ResNet-101.}
    \label{fig:catdet}
\end{figure}

We adopt CaTDet~\cite{mao2018catdet} as an example. CaTDet adds a tracker in the cascaded model to enable temporal feedback, which helps save the workload and improve accuracy. As shown in Figure~\ref{fig:catdet}, CaTDet models preserve the mAP well but substantially increases detection delay compared with other Faster R-CNN models. The CaTDet model with an internal confidence threshold of 0.01 achieves mAP of 0.555, which is very close to that of the Faster R-CNN model (0.561), however, it increases AD from 8.2 to 9.2. 

\begin{table}[b]
    \centering
    \begin{tabular}{l|ccc}
      & \multirow{2}{*}{R-FCN} & Faster  & DFF\\
      & & R-CNN &  R-FCN \\
    \hline
    \hline
    Mean & 33.5 & 17.8 & 43.3\\
    \hline
    Clipped Mean & 24.4 & 13.8 & 31.5\\
    \hline
    Off-Window  & \multirow{2}{*}{10.2\%} & \multirow{2}{*}{3.6\%} & \multirow{2}{*}{14.3\%}\\
    Percentage & & & \\
    \hline
    Expected Off-Window  & \multirow{2}{*}{5.3\%} & \multirow{2}{*}{0.4\%} & \multirow{2}{*}{10.2\%}\\
    Percentage & & & \\
    \end{tabular}
    \vspace{2mm}
    \caption{Statistics to show the heavy-tail effect of delay distribution: more than expected detections that exceed a 100-frame window. Clipped mean is the mean value computed with Equation~\ref{eq:delay}.} 
    \label{tab:tail_effects}
\end{table}


\section{Analysis of Delay}
\label{sec:analysis}

In this section, we analyze the characteristics of video object detection delay on the VIDT dataset and aim to provide our insights into the AD metric.

\subsection{Delay Distribution}

In Section~\ref{sec:metric}, we make an assumption that video object detection delay follows the discrete exponential distribution but with a heavy tail. Here we provide more examples and analysis to examine the actual distribution of delay.

\begin{figure}[t]
    \centering
    \includegraphics[width=0.52\textwidth]{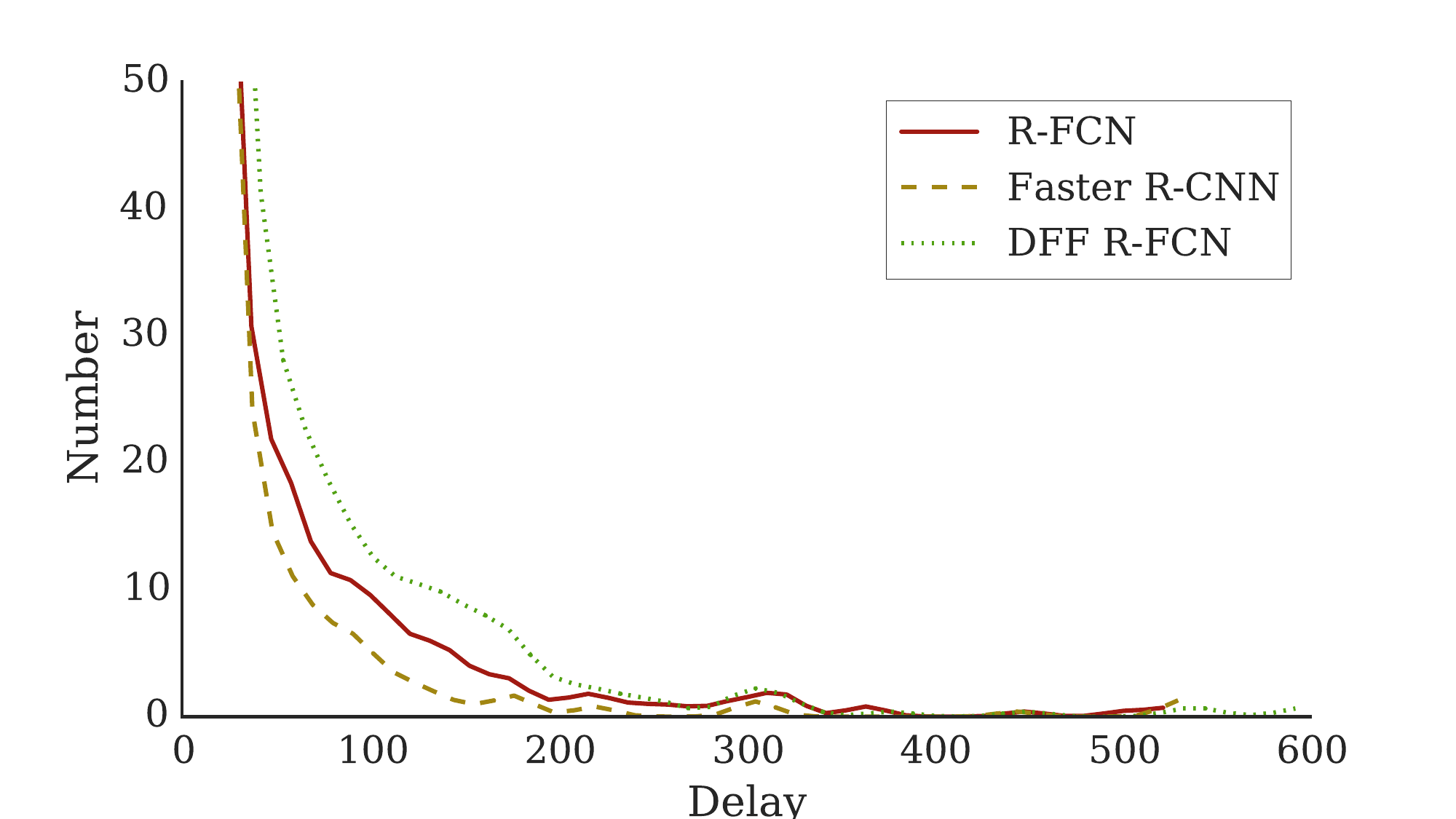}
    \caption{A zoomed-in plot of delay distribution of multiple detectors. All three models are based on ResNet-101 and have the same confidence threshold of 0.5. DFF runs with 1 key frame out of 10. }
    \label{fig:delay_multi}
\end{figure}

We select the three object detection methods: R-FCN, Faster R-CNN and DFF R-FCN, and plot their delay distribution in Figure~\ref{fig:delay_multi}. All three distributions resemble the exponential distribution. Note that at the same confidence threshold Faster R-CNN has the smallest delay, therefore its delay distribution is more skewed to left compared with the other two approaches. 

\begin{figure}[b]
    \centering
    \includegraphics[width=0.5\textwidth]{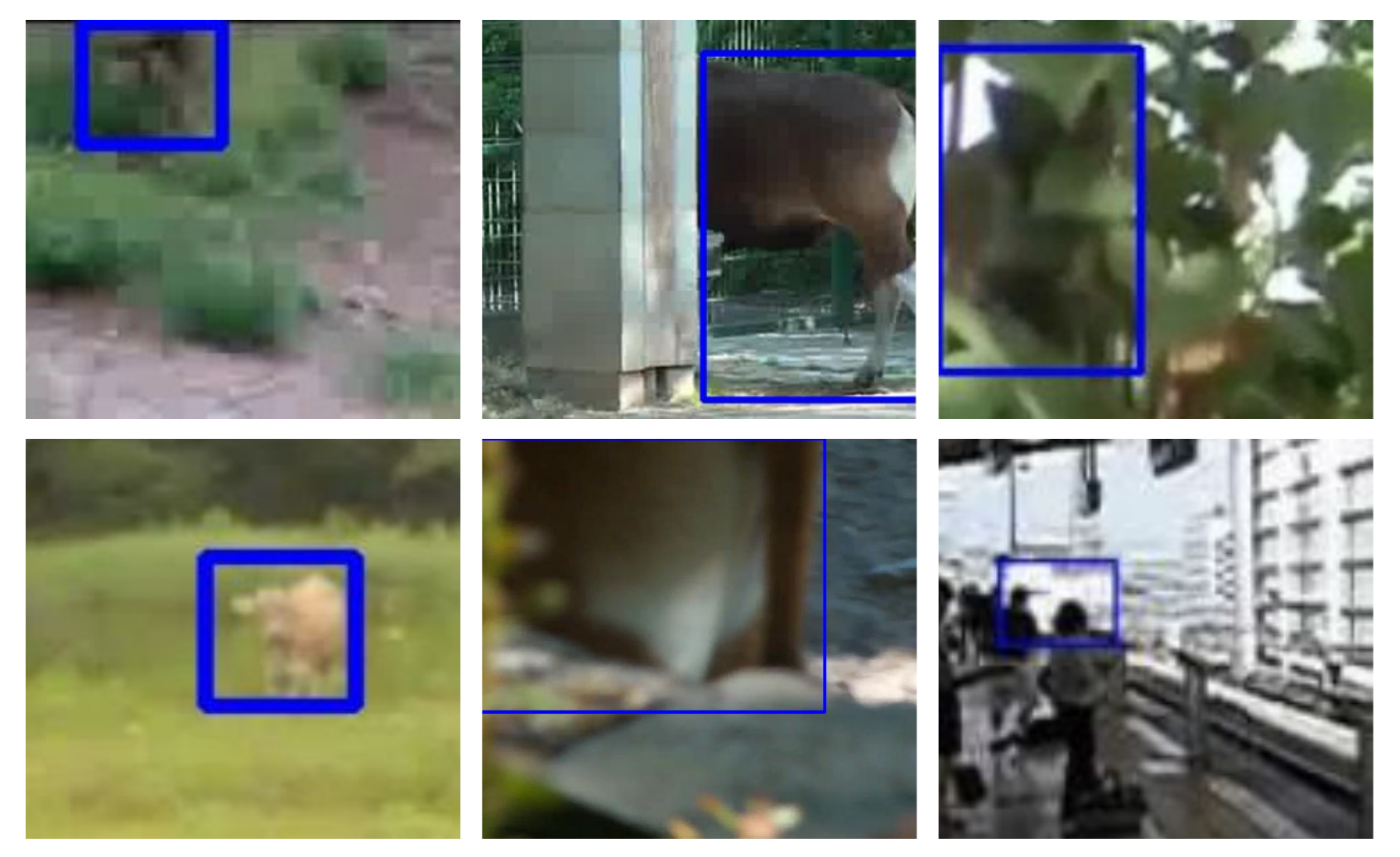}
    \caption{Examples of hard instances that have larger than 100-frame delay for R-FCN with ResNet-101. All crops are warped into the same dimensions. They represent some typical cases that tend to result in large detection delay: low resolution (left), severely occluded (mid), blurry and occluded (right).}
    \label{fig:VIDT_hard}
\end{figure}

We also show the statistics to measure the ``heavy-tail'' effect in Table~\ref{tab:tail_effects}. The difference between mean and clipped mean denotes that the long tail has a large impact on the mean value. Here we define ``expected off-window percentage'' as the probability of the delay $D$ falling out of a detection window, where $D$ is assumed to follow the ideal discrete exponential distribution. The ideal distribution is estimated by the maximum likelihood estimation. Such a probability can be computed by $P = (1-p)^W$, where $p$ is obtained as in Equation~\ref{eq:prob} and $W$ is the window size. The higher percentages outside the window in all three detectors validate that the tails are indeed ``heavier'' than those in the ideal exponential distributions. 
We select 6 examples out of 10 with largest delay and illustrate them in Figure~\ref{fig:VIDT_hard}. These video objects are either with very low resolution, heavily truncated or largely occluded.

\begin{figure}[t]
\centering
\includegraphics[width=0.53\textwidth]{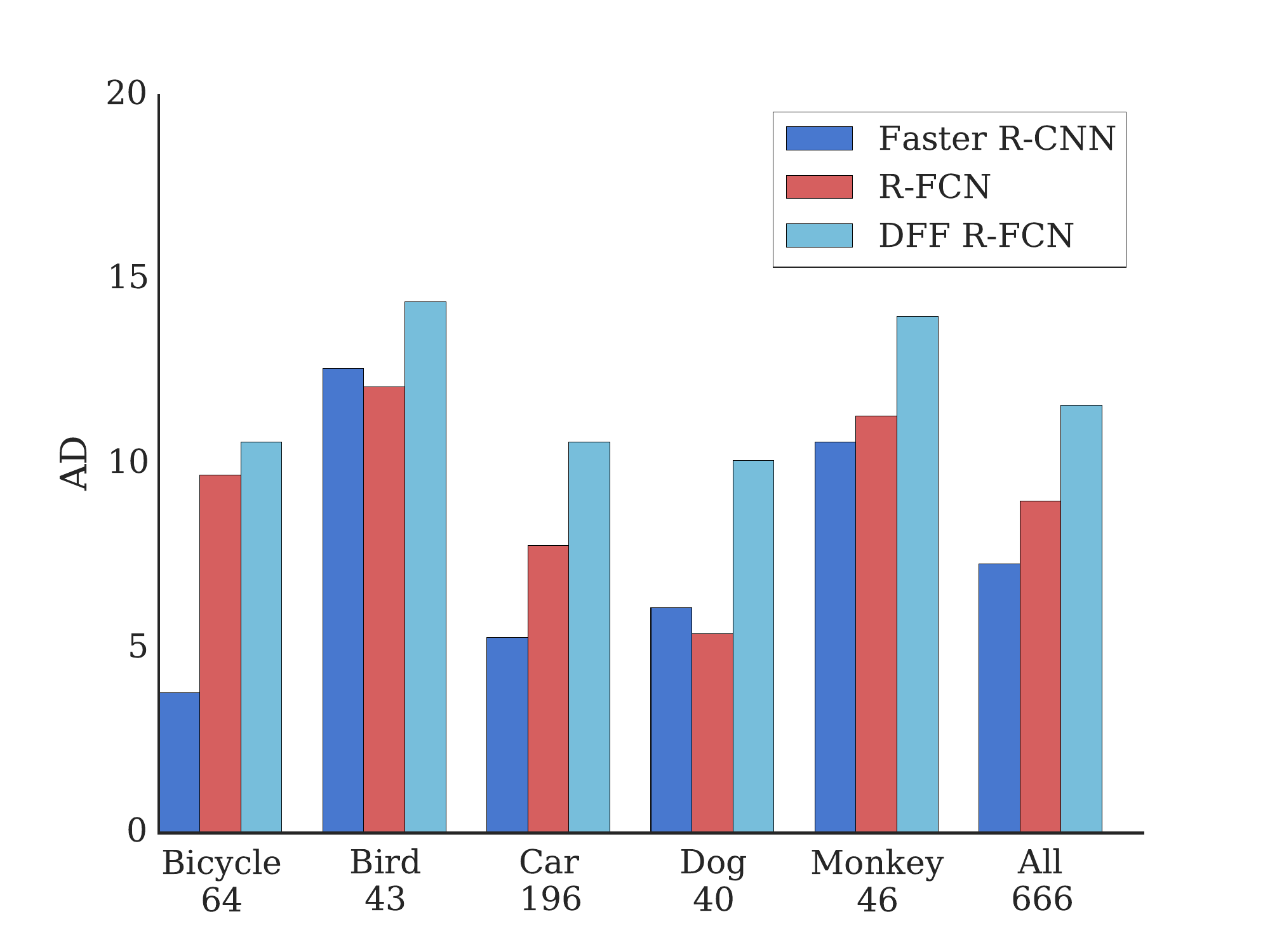}
\caption{AD by class: we only demonstrate the six video object classes, each of which contains more than 40 instances. The number of instances is shown under each class name. All three detectors adopt ResNet-101 as the base model. DFF operates with 1 key frame out of 10.}
\label{fig:ad_class}
\end{figure}

\subsection{Average Delay of Different Classes}

Due to the class imbalance in VIDT, AD is measured on all 666 instances instead of individual classes to avoid high variance. To demonstrate how the delay varies on different classes, we select 5 classes with over 40 instances and compare their AD results in Figure~\ref{fig:ad_class}.
All three models present large delays for class ``Bird'', which is typically small and quick moving. Classes ``Car'' and ``Dog'' have relatively smaller delays. For class ``Bicycle'', R-FCN and Faster R-CNN show distinct delays.  

\subsection{Average Delay of Different Scales}

To study how the size of instances affects detection delay, we divide all 666 instances into 3 categories by the averaged shorter dimension $D_s$ of their first 30 frames. \textit{Small}, \textit{Median} and \textit{Large} instances are categorized according to $D_s<40$, $40 \leq D_s < 100$ and $D_s \geq 100$. This criterion results in 129 small instances, 257 median instances and 280 large instances, respectively.

The anchor scale is the size of reference bounding box in all major object detection algorithms, where 3 and 4 scales are common choices for image object detection. As shown in Table~\ref{tab:scale_size}, further increasing the number of anchor scales from 3 to 4 does not improve mAP. However, adding a small scale helps with AD, in particular for the instances with lower resolutions. This is probably because that an instance is typically smaller when it appears in first few frames. The results with 5 scales show that further adding finer-grained scales does not help much.

\begin{table}[t]
    \centering
    \vspace{6pt}
\begin{tabular}{c|cccc|c}
Anchor  & \multirow{2}{*}{Small} & \multirow{2}{*}{Median} & \multirow{2}{*}{Large} & \multirow{2}{*}{Overall} & \multirow{2}{*}{mAP} \\
Scales & & & & & \\
\hline
\hline
 2  & 15.9  & 10.2   & 6.6   & 9.9     & 0.545 \\
 3  & 13.5  & 9.1    & 6.2   & 8.8     & 0.563 \\
 4  & \textbf{11.3}  & \textbf{9.0}      & \textbf{5.9}   & \textbf{8.2}     & 0.562 \\
 5  & 11.6  & 9.3    & 6     & 8.4     & \textbf{0.568} \\
\end{tabular}
\vspace{2mm}
\caption{Impact of anchor scales on AD for different instance sizes. The baseline model is Faster R-CNN with ResNet-50. 2 scales: (16, 32), 3 scales: (8, 16, 32), 4 scales: (4, 8, 16, 32), and 5 scales: (4, 6, 8, 16, 32).}
\label{tab:scale_size}
\end{table}

\subsection{Analysis of Variance}
Given the fact that VIDT only contains a few hundreds of instances, AD of various video object detectors evaluated on this dataset might be prone to high variance. Here we analyze if our comparisons are reliable, i.e., whether the difference between AD of different methods is significant compared to variance. 
To test the conclusion that DFF and FGFA incur extra delay to the baseline model R-FCN, we perform a 3-fold validation to verify whether the results correlate well on each fold. In addition, we select a subset from ImageNet VID-2017 (which is recently published but not yet widely used in the community) and validate whether the same conclusion can be extended to a different dataset. The results are shown in Table~\ref{tab:test_sig}. We find the results demonstrate good consistency across all folds and datasets. 

\begin{table}[t]
\begin{tabular}{c|cccc|c}
      & & \multicolumn{2}{l}{\hspace{0.7cm} VIDT}       &       & VIDT- \\
      & Fold 1 & Fold 2 & Fold 3 & Overall & 2017 \\
      \hline
      \hline
R-FCN  & 8.6     & 9.5     & 8.8     & 9.0       & 10.9                      \\
DFF & 8.7     & 10.1    & 9.1     & 9.2     & 11.0   \\  
FGFA & 9.3 & 11.5 & 10.1 & 10.2  & 12.2 \\      
\end{tabular}
\vspace{2mm}
\caption{Test of significance: AD results on different sub-folds of VIDT or another different dataset demonstrate good consistency. Here DFF runs with 1 key frame out of every 2 frames. }
\label{tab:test_sig}
\end{table}

\section{Conclusion}

We have presented the metric average delay (AD) to measure and compare detection delay of various video object detectors. Extensive experiments find that many detectors with descent detection accuracy suffer from the problem of increased delay.
However, the widely used detection accuracy metric mAP by itself cannot reveal this deficiency. 
We hope our findings and the new AD metric would help the design and evaluation of future video object detectors for latency-critical tasks. 
We also expect large and diverse video datasets in the future and better target the delay issue.

{\small
\bibliographystyle{ieee_fullname}
\bibliography{main}
}

\end{document}